# Classification of Fused Images using Radial Basis Function Neural Network for Human Face Recognition

Mrinal Kanti Bhowmik

Department of Computer Science and Engineering
Tripura University
Suryamaninagar- 799130, Tripura, India
e-mail: mkb_cse@yahoo.co.in

Debotosh Bhattacharjee, Mita Nasipuri, Dipak Kumar Basu*, Mahantapas Kundu

Department of Computer Science and Engineering
Jadavpur University
Kolkata- 700032, India
e-mail: debotosh@indiatimes.com,
mitanasipuri@gmail.com, dipakkbasu@gmail.com,
mkundu@cse.jdvu.ac.in
* AICTE Emeritus Fellow

*Abstract*—Here an efficient fusion technique for automatic face recognition has been presented. Fusion of visual and thermal images has been done to take the advantages of thermal images as well as visual images. By employing fusion a new image can be obtained, which provides the most detailed, reliable, and discriminating information. In this method fused images are generated using visual and thermal face images in the first step. In the second step, fused images are projected into eigenspace and finally classified using a radial basis function neural network. In the experiments Object Tracking and Classification Beyond Visible Spectrum (OTCBVS) database benchmark for thermal and visual face images have been used. Experimental results show that the proposed approach performs well in recognizing unknown individuals with a maximum success rate of 96%.

## I. INTRODUCTION

The subject of face recognition is as old as computer vision, both because of the practical importance of the topic and theoretical interest from cognitive scientists one of the most successful applications of image analysis and understanding. It is touch-less, highly automated and most natural since it coincides with the mode of recognition that any human being employs on his/her everyday affairs [1]. It has emerged as a preferred alternative to traditional forms of identification, like card IDs, which are not embedded into one's physical characteristics. Research into several biometric modalities including face, fingerprint, iris, and retina recognition has produced varying degrees of success [2]. Face recognition stands as the most appealing modality, since it is the natural mode of identification among humans and is totally unobtrusive. At the same time, however, it is one of the most challenging modalities [3]. It has many practical applications, such as bankcard identification, access control, mug shots searching, security monitoring, and surveillance systems [4], [5], [6].

Recently, researchers have investigated the use of fusion of thermal infrared and visual face images for person identification to tackle the drawbacks of individual thermal and visual images [7], [8], [9], [10], [11]. Despite the success of automatic face recognition techniques in many practical applications, the task of face recognition based only on the visible spectrum is still a challenging problem under uncontrolled environments. The challenges are even more profound when one considers the large variations in the visual stimulus due to illumination conditions, viewing directions or poses, facial expressions, aging, and disguises such as facial hair, glasses, or cosmetics. Out of these the major challenges are variations in illumination and pose [3]. Such problems are quite unavoidable in applications such as outdoor access control and surveillance. Performance of visual face recognition is sensitive to variations in illumination conditions and usually degrades significantly when the lighting is dim or when it is not uniformly illuminating the face. Illumination variation can cause changes in 2D appearance of an inherent 3D face object, and therefore, can seriously affect recognition performance. The changes caused by illumination are often larger than the differences between individuals. Various algorithms (e.g. histogram equalization, dropping leading eigenfaces etc.) for compensating such variations have been studied with partial success. These techniques attempt to reduce the within-class variability introduced by changes in illumination.

Thermal imaging [12] displays the amount of infrared energy emitted, transmitted, and reflected by an object. It has been suggested as a viable alternative in detecting disguised faces and handling situations where there is no control over illumination. Objects emit different amounts of IR energy according to their body temperature and characteristics. Since, vessels transport warm blood throughout the body; the thermal patterns of faces are derived primarily from the pattern of blood vessels under the skin. The vein and tissue structure of the face is unique for each person, and therefore the IR images are also unique. It is known that even identical twins have different thermal patterns. Anatomical features of faces useful for identification can be measured at a distance using passive IR sensor technology with or without the cooperation of the subject [3].

In this work at first thermal and visual face images are combined together and fused image of corresponding thermal and visual face images are obtained. After that using these transformed fused images eigenfaces are computed and finally those eigenfaces thus found are classified using a radial basis function neural network.

The organization of the rest of this paper is as follows. In section II, the overview of the system is discussed, in section III experimental results and discussions are given. Finally, section IV concludes this work.

## II. PREVIOUS WORK

The interest in human face detection analysis has been around for two decades and substantial efforts were made in the early 1990s. Over these years, researchers have focused on how to make face recognition systems fully automatic by tackling problems such as changes in illumination level and direction, changes in expression, changes in skin color due to cosmetics, glasses, beard, moustaches etc. However, a shortcoming of early works is their inability of performing automatic face detection. Meanwhile, significant advances have been made in the design of classifiers for successful face recognition. Among appearance-based holistic approaches, eigenfaces [13], [14], Fisherfaces [15], [16], [17] and interpolated bezier curve based face recognition [32] have proved to be effective in experiments with large databases. Feature-based graph matching approaches [18] have also been quite successful. Compared to holistic approaches, feature-based methods are less sensitive to variations in illumination and viewpoint and to inaccuracy in face localization. However, the feature extraction techniques needed for this type of approach are still not reliable or accurate enough [19]. But our experiments show better result than many other methods like Face Recognition using Gabor Filter, MLP, PCA for Visual indoor Probes etc.

## III. THE SYSTEM OVERVIEW

Here we present a technique for human face recognition. In this work we have used Object Tracking and Classification Beyond Visible Spectrum (OTCBVS) database benchmark thermal and visual face images. Every thermal face image and visual face image is first combined and converted into fused image. These transformed images are separated into two groups namely training set and testing set. The eigenspace is computed using training images. All the training and testing images are projected into the created eigenspace and named as fused eigenfaces. Once these conversions are done the next task is to use a classifier to classify them. A multilayer feed forward network is used for this purpose. The block diagram of the system is given in Figure 1. In this figure dotted line indicates feedback from different steps to their previous steps to improve the efficiency of the system e.g. if the classification results are not satisfactory some adjustment in the internal parameters may be done. In case of eigenspace projection number of eigenvectors, used to create eigenspace, may be increased or decreased to improve better representation of eigenfaces in the reduced space in order to achieve higher accuracy in recognition of faces.

### A. Visual Face Images

There are few problems regarding visual face recognition in case of uncontrolled operating environments such as outdoor situations and low illumination conditions. Visual face recognition also has difficulty in detecting disguised faces, which is critical for high-end security applications, so for that reason thermal face recognition came into picture.

### B. Thermal Infrared Face Images

Thermal infrared face images are formed as a map of the major blood vessels present in the face. Therefore, a face recognition system designed based on thermal infrared face images cannot be evaded or fooled by forgery, or disguise, as can occur using the visible spectrum for facial recognition. Compared to visual face-recognition systems this recognition system will be less vulnerable to varying conditions, such as head angle, expression, or lighting.

### C. Image Fusion Technique

Image fusion is the process by which two or more images are combined into a single image retaining the important features from each of the original images.

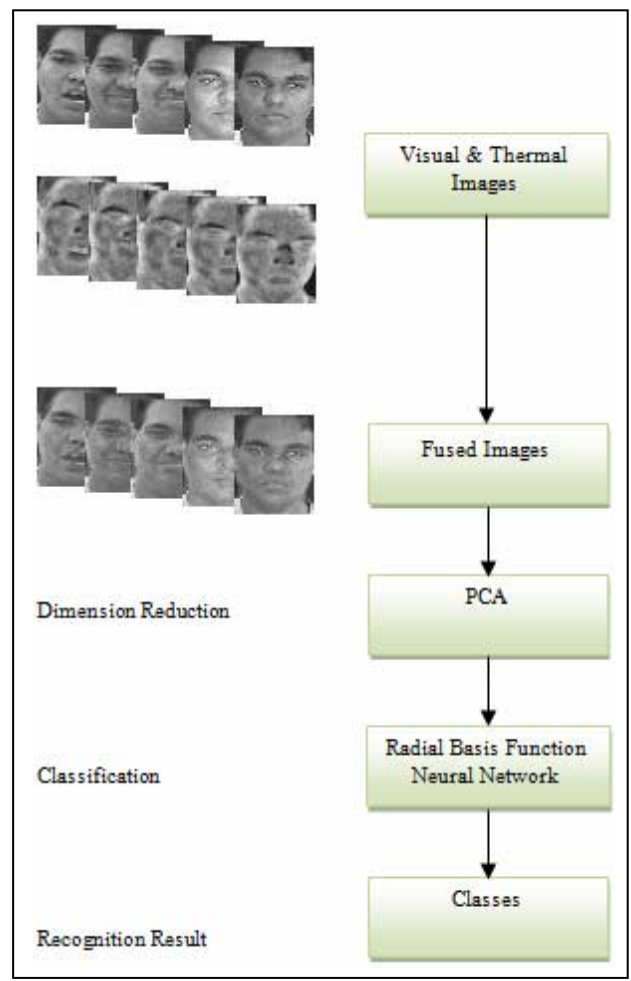

Figure 1. Block diagram of the system presented here.

The fusion of images is often required for images acquired from different instrument modalities or capture techniques of the same scene or objects. Several approaches to image fusion can be distinguished, depending on whether the images are fused in the spatial domain or they are transformed into another domain, and their transforms fused.

The interpretation of each type of image thus leads to ambiguous inferences about the nature of the objects in the scene. The use of thermal data gathered by an infrared camera, along with the visual image, is seen as a way of resolving some of these ambiguities. In the other hand, thermal images are obtained by sensing radiation in the infrared spectrum. The radiation sensed is either emitted by an object at a non-zero absolute temperature, or reflected by it. The mechanisms that produce thermal images are different from those that produce visual images. Thermal image produced by an object's surface can be interpreted to identify these mechanisms. Thus thermal images can provide information about the object being imaged which is not available from a visual image [11].

Some work has been done in recognizing objects in a scene using infrared images. However, there has been little effort on interpreting thermal images of outdoor scenes based on a study of the mechanism that gives rise to the differences in the thermal behavior of object surfaces in the scene. Also, nor has been any effort been made to integrate information extracted from the two modalities of imaging.

Here, 70% of visual image and 30% of thermal image of same class or same image have been brought together into a common operating image and that is commonly referred to as a Common Relevant Operating Picture (CROP) [10]. Thus, the degree of filtering and intelligence applied to the pixel streams is increased to present pertinent information to the user. So image pixel fusion has the capacity to enable seamless working in a heterogeneous work environment with more complex data. For accurate and effective face recognition we require more informative images. Combining the features of both the images viz. visual and thermal face images is efficient, robust, and accurate for face recognition because one source (i.e. thermal) may lack some information which might be available in images by other source (i.e. visual).

The fusion scheme considered in this work is describe below. Here, pair of images, one in the IR spectrum and one in the visible spectrum represents each face. Both images have been combined prior to fusion to ensure similar ranges of values.

We fused visual and thermal images. Ideally, the fusion of common pixels can be done by pixel-wise weighted summation of visual and thermal images[20], given below:

$$F(x, y) = a(x, y)V(x, y) + b(x, y)T(x, y) \qquad (1)$$

where $F(x, y)$ is a fused output of a visual image, $V(x, y)$, and a thermal image, $T(x, y)$, while $a(x, y)$ and $b(x, y)$ represent the weighting factors for visual and thermal images respectively. In this work, we have considered $a(x, y) = 0.70$ and $b(x, y) = 0.30$.

*D. Eigenfaces for Recognition*

In mathematical terms, we wish to find principal components [14] [21] [22] of the distribution of faces, or the eigenvectors of the covariance matrix of the set of face images.

These eigenvectors can be thought of as set of features which together characterize the variations between face images. Each image location contributes more or less to each eigenvector, so that we can display the eigenvector as sort of ghostly face which we call an eigenface. Each face image in the training set can be presented exactly in terms of a linear combination of the eigenfaces. The number of a possible eigenfaces is equal to the number of face images in the training set. However the faces can also be approximated using only the "best" eigenfaces-those that have the largest eigenvalues, and which therefore account for the most variance within the set face images. The best U eigenfaces constitute a U-dimensional subspace, which may be called as "face space" of all possible images. Identifying images through eigenspace projection takes three basic steps. First the eigenspace must be created using training images. After that all those training images are projected into the eigenspace and call them eigenfaces. Finally, the test images are identified by projecting them into the eigenspace and classifying them by the trained classifier.

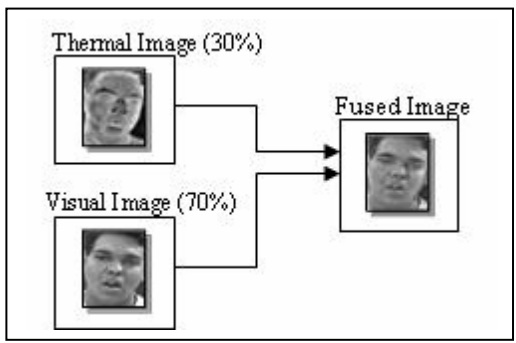

Figure 2. Fusion Technique.

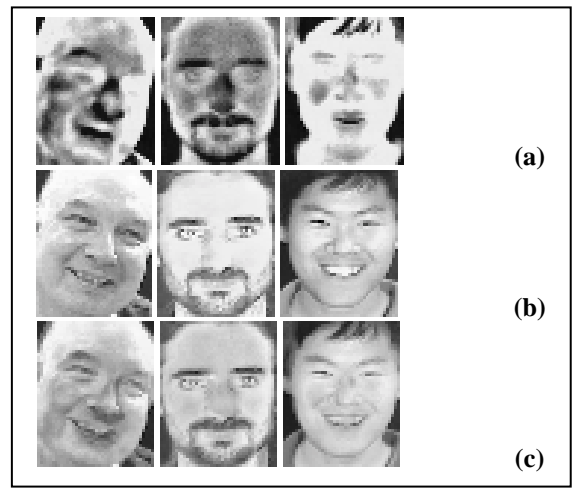

Figure 3. (a) Thermal Images, (b) Visual Images, (c) Fused Images of corresponding thermal and visual images.

*E. Classification of Fused Eigenfaces using Radial Basis Function Network [23]*

Neural networks have been employed and compared to conventional classifiers for a number of classification problems. The results have shown that the accuracy of the neural network approaches is equivalent to, or slightly better than, other methods. Also, due to the simplicity, generality and good learning ability of the neural networks, these types of classifiers are found to be more efficient.

Radial Basis Function (RBF) neural networks are found to be very attractive for many engineering problems because (1) they are universal approximates, (2) they have a very compact topology and (3) their learning speed is very fast because of their locally tuned neurons. An important property of RBF neural networks is that they form a unifying link between many different research fields such as function approximation, regularization, noisy interpolation and pattern recognition. Therefore, RBF neural networks serve as an excellent candidate for pattern classification where attempts have been carried out to make the learning process in this type of classification faster than normally required for the multilayer feed forward neural networks [24].

In this paper, an RBF neural network is used as a classifier in a face recognition system where the inputs to the neural network are feature vectors derived from the proposed feature extraction technique described in II-B.

Geometrically, the key idea of an RBF neural network is to partition the input space into a number of subspaces which are in the form of hyperspheres. Accordingly, clustering algorithms (k-means clustering, fuzzy k-means clustering and hierarchical clustering) which are widely used in RBF

## IV. EXPERIMENTAL RESULTS AND DISCUSSIONS

All the experiments have been done using MATLAB 7. To covers all conditions of human face recognition, a thorough system performance investigation has been conducted. They are face recognition under i) variations in size, ii) variations in lighting conditions, iii) variations in facial expressions, iv) variations in pose.

We first analyze the performance of our algorithm using OTCBVS database which is a standard benchmark thermal and visual face images for face recognition technologies.

*A. OTCBVS Database*

Object Tracking and Classification Beyond Visible spectrum (OTCBVS) benchmark database contains 700 images for each of the visual and thermal images of 16 different persons. The images were taken at different times which contain quite a high degree of variability in lighting, facial expression (open / closed eyes, smiling / non smiling etc.), pose (Up right, frontal position etc.) and facial details (Glasses/ no Glasses). All the images were taken against a dark homogeneous background with the subjects in and upright, fontal position, with tolerance for some tilting and rotation of up to 20 degree. The variation in scale is up to about 10% all the images in the database.

*B. Classification of Fused Eigenfaces using Radial Basis Function Neural Network*

Out of total 700 thermal and visual images 400 images are taken out of which 200 are thermal images and 200 are visual images. Combining these thermal and visual images we get 200 fused images. 100 of these images are used as training set and rest 100 images are taken as testing images. The training set contains 10 classes which mean that each class has 10 images. Now 5 images from one particular class (which are not used as a training image) and 5 more images of the other classes are taken from the testing set. According to this process for all the 10 classes we get the results which are shown in the following Table 1.

TABLE I. STUDY OF TESTING PHASE OF FUSED IMAGES

| Class | Total number of testing images | Number of testing images from one particular class | Number of testing images from other 5 different classes | Recognition rate | False rejection rate |
|---|---|---|---|---|---|
| Class-1 | 10 | 5 | 5 | 86% | 14% |
| Class-2 | 10 | 5 | 5 | 88% | 12% |
| Class-3 | 10 | 5 | 5 | 79% | 21% |
| Class-4 | 10 | 5 | 5 | 87% | 13% |
| Class-5 | 10 | 5 | 5 | 96% | 5% |
| Class-6 | 10 | 5 | 5 | 82% | 18% |
| Class-7 | 10 | 5 | 5 | 86% | 14% |
| Class-8 | 10 | 5 | 5 | 79% | 21% |
| Class-9 | 10 | 5 | 5 | 83% | 17% |
| Class-10 | 10 | 5 | 5 | 93% | 7% |

In the following figures (from Fig. 4 to Fig. 9) fused images have been shown, which were used in the testing set of class 1, 2, and 3 with different expressions, variations and the different other conditions (lightening, darkness etc.) are mentioned with each of the fused face images.

Different variations of 10(5+5) numbers of testing images (Training is done by 100 images; from these 100 images 10 images of class-1 are used):

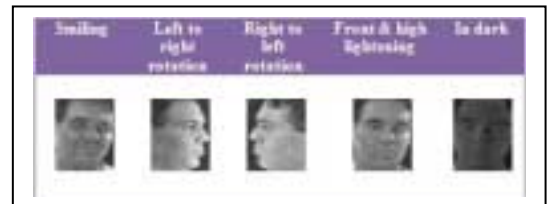

Figure 4. Fused images of class-1 for testing (which are not used in training).

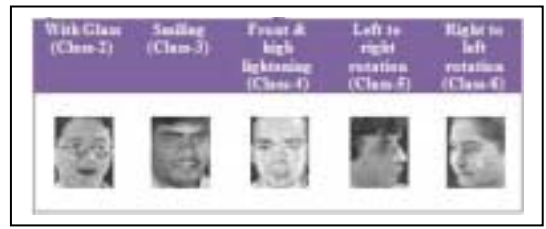

Figure 5. Fused images of other classes except class-1 (which are not used in training).

Different variations of 10 (5+5) numbers of testing images (Training is done by 100 images; from these 100 images 10 images of class-2 were used):

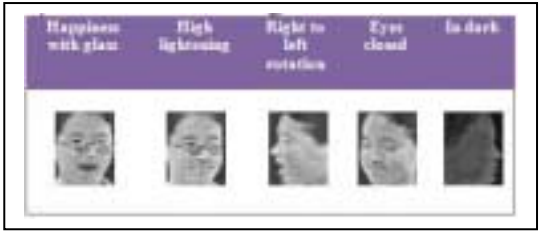

Figure 6.  Fused images of class-2 for testing (which were not used in training).

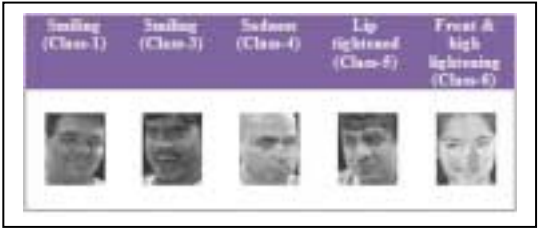

Figure 7.  Fused images of other classes except class-2 (which were not used in training).

Different variations of 10 (5+5) numbers of testing images (Training is done by 100 images; from these 100 images 10 images of class-3 were used):

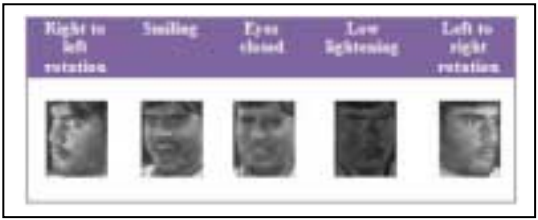

Figure 8.  Fused images of class-3 for testing (which were not used in training).

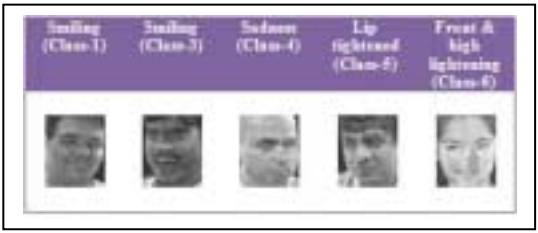

Figure 9.  Fused images of other classes except class-3 (which were not used in training).

## C. Comparison with Other Methods

TABLE II.  COMPARISON OF RECOGNITION RATE WITH OTHER METHODS

| Method | Recognition rate |
|---|---|
| Present Method | 85.9% (96% maximum) |
| PCA [28] | 83.4% |
| PCA for Visual indoor Probes[29] | 81.54% |
| Wavelet Subband + Kernel associative Memory with XM2VTS database [30] | 84% |
| Face Recognition using Gabor Filter [31] | 85% |

Comparison of recognition rates for the present method and other commonly referred methods are shown in Table 5 for a quick comparison. Although they use different face databases, i.e. other than OTCBVS face database, the present method can be compared favorably against other face recognition methods.

## V. CONCLUSION

In this paper we have presented an image pixel based fusion technique for face recognition. After the fusion of images as weighted sum, the fused images are projected into eigenspace. Those fused eigenfaces are classified using radial basis function neural network. The efficiency of our scheme has been demonstrated on Object Tracking and Classification Beyond Visible spectrum (OTCBVS) benchmark database and recognition rate obtained is 96 %.

ACKNOWLEDGMENT

Author, M. K. Bhowmik is thankful to the project entitled "Development of Techniques for Human Face Based Online Authentication System Phase-I" sponsored by Department of Information Technology under the Ministry of Communications and Information Technology, New Delhi-110003, Government of India Vide No. 12(14)/08-ESD, Dated 27/01/2009 at the Department of Computer Science & Engineering, Tripura University-799130, Tripura (West), India for providing the necessary infrastructural facilities for carrying out this work.